\documentclass{article}

 \usepackage{wrapfig}

\usepackage[main, final]{neurips_2025}

\usepackage[utf8]{inputenc} 
\usepackage[T1]{fontenc}    
\usepackage{hyperref}       
\usepackage{url}            
\usepackage{booktabs}       
\usepackage{amsfonts}       
\usepackage{nicefrac}       
\usepackage{microtype}      
\usepackage{xcolor}         

\usepackage{multirow}
\usepackage{colortbl}
\usepackage{makecell}
\usepackage{amssymb}
\usepackage{gensymb}
\definecolor{citecolor}{HTML}{0071BC}
\definecolor{linkcolor}{HTML}{ED1C24}
\definecolor{deemph}{gray}{0.6}
\usepackage{tikz}
\usepackage{amsmath}
\usepackage{float}
\newlength\savewidth

\let\For\relax
\let\State\relax
\usepackage[ruled,vlined]{algorithm2e}
\newcommand{\NoLabelLine}[1]{%
  \SetAlgoNoLine      
  \noindent #1\par
  \SetAlgoVlined
}

\usepackage{titlesec}
\titlespacing\section{0pt}{0pt plus 0pt minus 0pt}{0pt plus 0pt minus 0pt}
\titlespacing\subsection{0pt}{0pt plus 0pt minus 0pt}{0pt plus 0pt minus 0pt}
\titlespacing\subsubsection{0pt}{0pt plus 0pt minus 0pt}{0pt plus 0pt minus 0pt}

\usepackage{fontawesome}
\title{Raw2Drive: Reinforcement Learning with Aligned World Models for End-to-End Autonomous Driving\\(in CARLA v2)}

\definecolor{darkblue}{rgb}{0, 0, 0.5}
\hypersetup{colorlinks=true, citecolor=darkblue, linkcolor=darkblue, urlcolor=darkblue}

\author{%
  Zhenjie Yang$^{1}$, Xiaosong Jia$^{2\dagger}$, Qifeng Li$^{1}$, Xue Yang$^1$, Maoqing Yao$^3$, Junchi Yan$^{1\dagger}$ \\ \\
  1. Sch. of CS, Sch. of AIS, Sch. of AI, Shanghai Jiao Tong University \\
  2. Institute of Trustworthy Embodied AI, Fudan University \quad
  3. AgiBot \\
}

\begin{document}

\maketitle
\begin{abstract}
Reinforcement Learning (RL) can mitigate the causal confusion and distribution shift inherent in imitation learning (IL). However, applying RL to end-to-end autonomous driving (E2E-AD) remains an open problem for its training difficulty, and IL is still the mainstream paradigm in both academia and industry. Recently Model-based Reinforcement Learning (MBRL) have demonstrated promising results in neural planning; however, these methods typically require privileged information as input rather than raw sensor data. We fill this gap by designing \textbf{\textit{Raw2Drive}}, a dual-stream MBRL approach. Initially, we efficiently train an auxiliary privileged world model paired with a neural planner that uses privileged information as input. Subsequently, we introduce a raw sensor world model trained via our proposed \textbf{\textit{Guidance Mechanism}}, which ensures consistency between the raw sensor world model and the privileged world model during rollouts. Finally, the raw sensor world model combines the prior knowledge embedded in the heads of the privileged world model to effectively guide the training of the raw sensor policy. Raw2Drive is so far the only RL based end-to-end method on CARLA Leaderboard 2.0, and Bench2Drive and it achieves state-of-the-art performance.
\end{abstract}

\section{Introduction}
\renewcommand{\thefootnote}{} 
\footnotetext{\quad $^\dagger$ Correspondence Author. This work was in part supported by by NSFC (62222607,62506229) and Natural Science Foundation of Shanghai (25ZR1402268).} 
\renewcommand{\thefootnote}{\arabic{footnote}}
\label{sec:intro}

\begin{figure}[]
    \centering
    \includegraphics[width=1\linewidth]{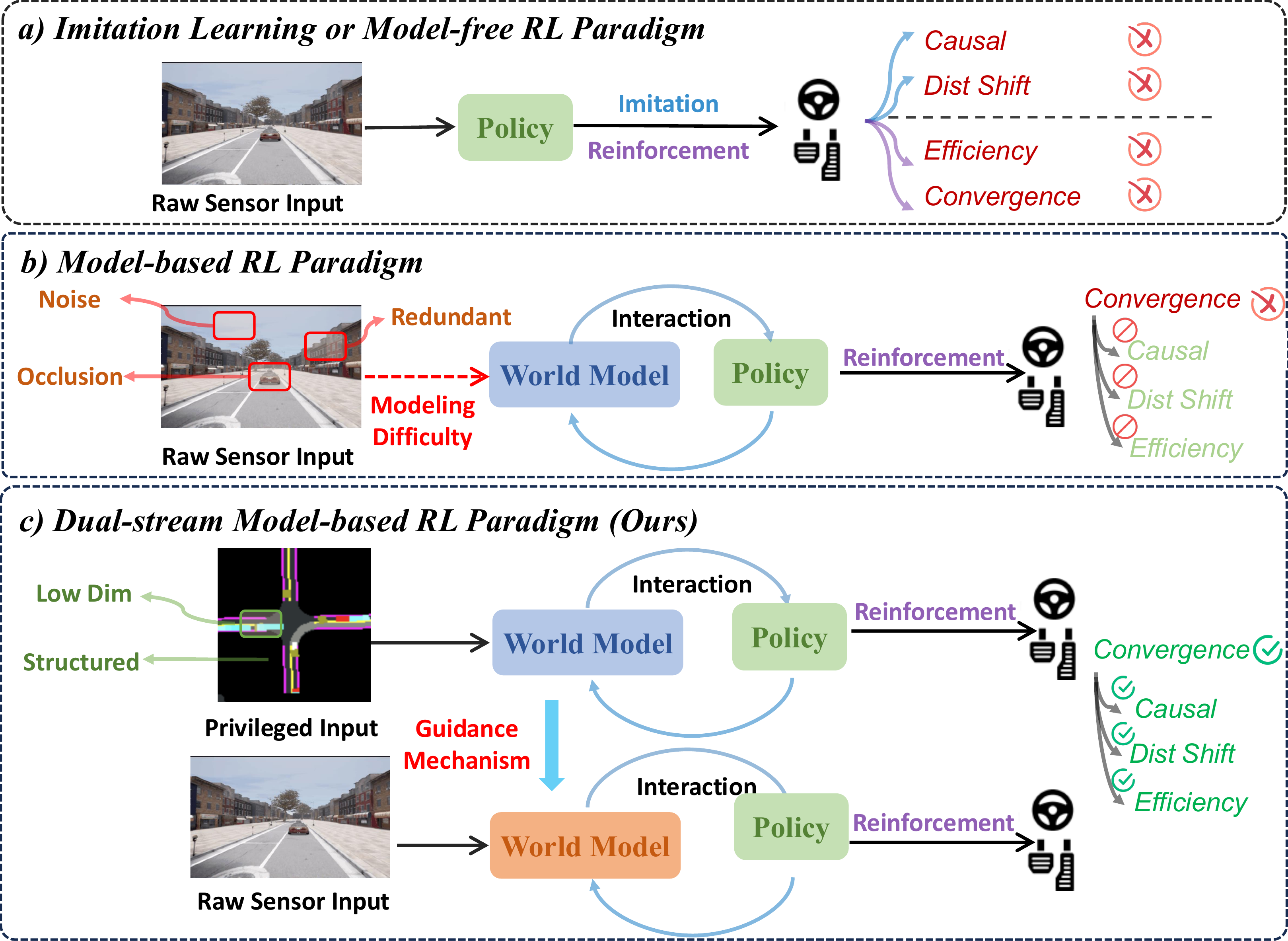}
    \caption{\textbf{Comparison of different training paradigms in end-to-end autonomous driving.} \textbf{(a) Imitation Learning} suffers from causal confusion~\cite{d2014confusion} and distribution shift~\cite{wen2020fighting}. \textbf{Model-free Reinforcement Learning}~\cite{toromanoff2020end} faces efficiency problem and fails to converge. \textbf{(b) Model-based Reinforcement Learning}: There are no reported such works for raw sensor input E2E-AD as the raw data can be noisy and redundant, and Think2Drive~\cite{li2024think} assumes the privileged ground truth data is given, which cannot be directly applied in real-world AD. \textbf{(c)} In \textbf{\textit{Raw2Drive}}, we propose the first feasible model-based reinforcement learning paradigm for end-to-end autonomous driving. By leveraging low-dimensional, structured privileged input, our approach guides the learning of a world model from raw sensor data, effectively addressing the issues outlined in (a) and (b).
    \label{fig:paradigm}}
\end{figure}

Beyond modular systems, end-to-end autonomous driving models~\cite{chen2024end, Chitta2023PAMI, wu2023policypretrainingautonomousdriving, li2023delving} are emerging, where a unified model directly uses raw sensor inputs for planning. As shown in Figure~\ref{fig:paradigm}, most of these models~\cite{hu2023planning, jiang2023vad, jia2022multi, jia2023towards} are based on imitation learning (IL), which trains models to mimic expert demonstrations. However, \textbf{IL faces fundamental limitations such as poor generalization to unseen situations~\cite{wen2020fighting, zhai2023ADMLP, wu2022trajectoryguided, jia2021ide} and causal confusion~\cite{d2014confusion, jia2023thinktwice, jia2023driveadapter}}, which occur when the model incorrectly associates actions with the wrong causes. These issues become problematic in complex and dynamic driving scenarios where decision-making must account for delicate interactions with environment.

Reinforcement learning (RL)~\cite{mnih2015human, niu2023lightzero, pu2024unizero} offers a promising alternative by optimizing driving policies through reward-driven interactions with the dynamic environment~\cite{schulman2017proximal, mnih2013playing, Bai_Liu_Du_Wen_Yang_2025}, beyond the expert demonstration in imitation learning. Recently, RL has been shown to achieve higher performance bounds than IL in various domains, such as  AlphaZero~\cite{silver2017mastering, niu2024lightzero}, and OpenAI-O3/DeepSeek-R1~\cite{guo2025deepseek}. These successes highlight \textbf{RL’s ability to adapt policies for complex decision-making tasks.}

As shown in Table~\ref{tab:compare}, compared to the extensive research in IL for AD, either through modular systems~\cite{Mobileye, NVIDIA} or recent end-to-end approaches~\cite{hu2023planning}, RL for AD has been comparatively less explored. Pioneering work MaRLn~\cite{toromanoff2020end} faced significant efficiency and convergence issues, requiring approximately 50M steps (57 days of training) with performance lagging far behind contemporary IL based methods~\cite{chen2020learning,Prakash2021CVPR}. 
As of today, IL-based methods achieve saturated performance on CARLA Leaderboard 1.0~\cite{carlaStartedWith1} (abbreviated as CARLA v1), while \textbf{none of them can achieve satisfying scores on the more challenging CARLA Leaderboard 2.0}~\cite{carlaStartedWith2} (abbreviated as CARLA v2).

This situation recently changed with the seminal work Think2Drive~\cite{li2024think}, which introduced a model-based RL (MBRL) approach that successfully solved CARLA v2, with the help of the world model~\cite{hafner2020dreamer}. However,  Think2Drive~\cite{li2024think} relies on privileged information (ground-truth states of environments) and \textbf{there is still no success reported in the field to apply MBRL with raw sensor inputs}. The main challenge lies in the contrast between privileged information,  which is compact and facilitates efficient training, and raw sensor data, which is high-dimensional, redundant, and noisy, making the training of world model significantly more difficult. As shown in Table~\ref{tab:compare}, there are no RL based end-to-end methods in CARLA v2, which is the very setting that this paper seeks to challenge. Due to limited space, we discuss more details of \textbf{Related Works} in Appendix~\ref{sup:related_work}.

\begin{table}[]
\caption{\textbf{Comparison of settings of mainstream algorithms}. WM means whether to use the world model. SSL means self-supervised learning. CoRL2017~\cite{dosovitskiy2017carla} and CARLA v1~\cite{carlaStartedWith1} include 4 and 10 standard cases, respectively. CARLA v2~\cite{carlaStartedWith2, jia2024bench}, on the other hand, introduces 39 additional real-world corner cases, which are significantly more difficult. ROM03 is based on CARLA v1 including 4 very basic scenes.}
\label{tab:compare}
\centering
\resizebox{1\linewidth}{!}{
\begin{tabular}{l|c|c|c|c|c|c|c}
\toprule
Method                           & Venue                          & Input     & Scheme & E2E & w/ WM & Benchmark & Corner Case \\ \hline
Chen~\cite{chen2019model}            & ITSC 2019 & Privileged & RL       & No         & No          & Roundabout & No      \\ \hline
MaRLn~\cite{toromanoff2020end}      & CVPR 2020 & Raw       & RL       & Yes        & No          & CoRL 2017    & No      \\ \hline
Roach~\cite{zhang2021end}            & ICCV 2021 & Privileged & RL       & No         & No          & Carla v1     & No      \\ \hline
UniAD~\cite{hu2023planning}          & CVPR 2023 & Raw       & IL       & Yes        & No          & Carla v1     & No      \\ \hline
Think2Drive~\cite{li2024think} & ECCV 2024& Raw       & RL       & No         & Yes         & Carla v2     & Yes     \\ \hline
LAW~\cite{li2025enhancing}           & ICLR 2025 & Raw       & IL+SSL      & Yes        & Yes          & Carla v1     & No      \\ \hline
AdaWM~\cite{wang2025adawmadaptiveworldmodel} & ICLR 2025 & Privileged & RL & No & Yes & ROM03 & No  \\ \hline
DriveTrans~\cite{jia2025drivetransformer}  & ICLR 2025 & Raw       & IL       & Yes        & No          & Carla v2     & Yes     \\ \hline
Raw2Drive(Ours)     & - & Raw       & RL       & Yes        & Yes         & Carla v2     & Yes     \\ \bottomrule
\end{tabular}
}
\end{table}

In this work, we propose \textbf{\textit{Raw2Drive}}, a dual-stream MBRL method, achieves state-of-the-art performance on CARLA v2 and surpasses IL based methods by a large margin. The training is divided into two stages. In the first stage, we leverage the privileged information to train a privileged world model and a paired neural planner. Then, we further jointly train a raw sensor world model and an end-to-end planner whose input is directly the raw video. In the training of the raw sensor world model, instead of reconstructing computationally expensive multi-view videos, it is guided by the alignment with the frame-wise feature of the trained privileged world model, facilitating an otherwise extremely difficult training process. We further propose the \textbf{\textit{Guidance Mechanism}} to enforce alignment, ensuring consistency in future state predictions during rollouts across both world models. Additionally, it leverages the prior knowledge embedded in the heads of the privileged world model to effectively guide the training of the raw sensor policy. \textbf{The contributions are as follows:}

\noindent\textbf{•}\quad To our best knowledge, Raw2Drive is the first MBRL framework for E2E-AD, i.e. from raw image input to planning, beyond existing IL~\cite{hu2023planning,Chitta2023PAMI} or privileged input based RL approaches~\cite{zhang2021end, li2024think}.
    
\noindent\textbf{•}\quad Raw2Drive achieves state-of-the-art performance on the challenging CARLA v2 and Bench2Drive and surpasses IL methods by a large margin, validating the power of RL.

\noindent\textbf{•}\quad We only use 64 H800 GPU days in total to deliver our final planner, and the cost can be further saved to 40 GPU days when Think2Drive is reused which dismisses our phase I training. In comparison, IL-based UniAD costs about 30 GPU days yet it only solves even 3$\sim$4 corner cases in CARLA v2. We believe it is significantly less than the imitation learning approaches used in industry and the hope is our work could provide an orthogonal reference for industry applications. 

\begin{figure*}[thb!]
    \centering
    \includegraphics[width=1\linewidth]{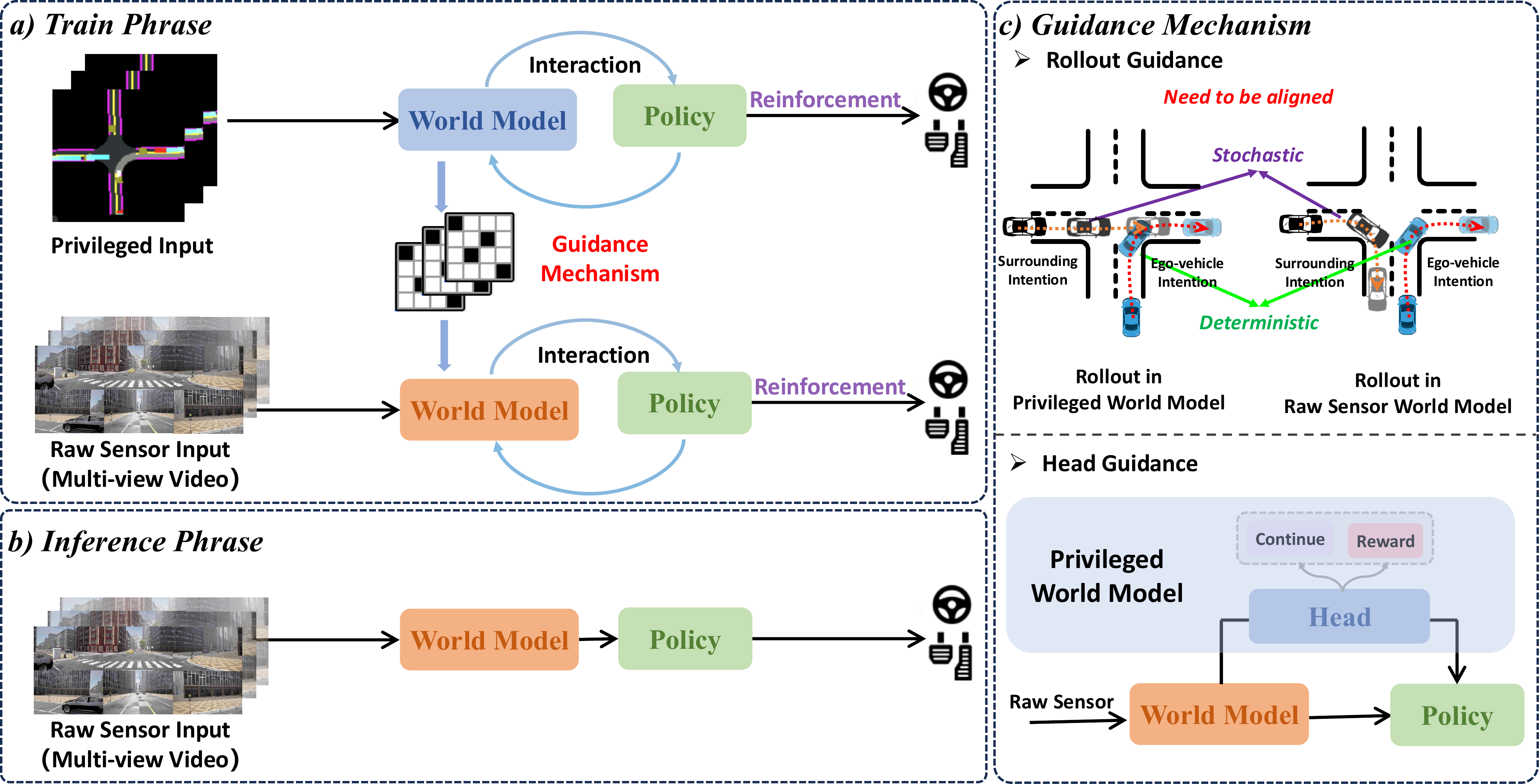}
    \caption{\textbf{The Overall Pipeline of Raw2Drive.} (a) During training, we use privileged input to train the privileged world model and paired policy. Then, the privileged world model is used to guide the training of the raw sensor stream. (b) During inference, only raw sensor inputs are available, which aligns with real-world autonomous driving. (c) The guidance mechanism consists of two parts: (I) Rollout Guidance to ensure future modeling consistency; (II) Head Guidance to ensure the supervision for raw sensor policy is accurate and stable.}
    \label{fig:method}
\end{figure*}

\section{Problem Formulation \& Related Works}
\label{sec:problem_setup}
In Figure~\ref{fig:method}, we give a problem formulation where there are two streams: (I) privileged observations $o_t$ with ground-truth bounding boxes, and HD-Map. (II) raw sensor observations $\hat{o}_t$ with onboard sensors such as cameras, LiDAR, and IMU. Note that \textbf{privileged observations are only accessible during training} to facilitate learning. During inference, only raw sensor observations are allowed.

\section{Method}
\label{sec:method}

\begin{table*}[tb!]
\caption{\textbf{Notations of Dual Stream World Models in \textit{Raw2Drive}}. $\text{WM}$ means the privileged world model  and  $\hat{\text{WM}}$ means the raw sensor world model. $t$ denotes the time-step. Both world models consist of an Encoder, RSSM, and Heads. The privileged world model is similar to DreamerV3~\cite{hafner2023mastering} while raw sensor world model has a tailored encoder $\hat{Enc}$ and only has a tailored decoder head $\hat{Dec}$. $\textcircled{1}$ and $\textcircled{2}$ represent the operation during training and inference respectively.} 
\label{table:notation}
\centering
\resizebox{1\linewidth}{!}{
\begin{tabular}{cccc|c|c}
\toprule
\multicolumn{4}{c|}{Dual Stream} &
  Privileged ${\text{WM}}$ &
  Raw Sensor ${\hat{\text{WM}}}$ \\ \hline
\multicolumn{4}{c|}{Observation} &
  $o_{t}$ &
  $\hat{o}_{t}$ \\ \hline
\multicolumn{4}{c|}{\text{Encoder State}} &
  $e_{t}=\text{Enc}(o_{t})$ &
  $\hat{e}_{t}=\hat{\text{Enc}}(\hat{o}_{t})$ \\ \hline
\multicolumn{1}{c|}{\multirow{6}{*}{\begin{tabular}[c]{@{}c@{}}World\\ Model\end{tabular}}} &
  \multicolumn{1}{c|}{\multirow{3}{*}{RSSM}} &
  \multicolumn{2}{c|}{Deterministic State} &
  $h_{t}=f_\theta(h_{t-1},a_{t-1},s_{t-1})$ &
  \begin{tabular}[c]{@{}c@{}}$\hat{h}_{t}=\hat{f}_\theta(\hat{h}_{t-1},\color{black}a_{t-1}\color{black}, \color{black}s_{t-1}\color{black})~\tikz[baseline]{\node[draw=black,fill=white,text=black,circle,inner sep=1pt,scale=0.7] {1};}$\\ $\hat{h}_{t}=\hat{f}_\theta(\hat{h}_{t-1},\hat{a}_{t-1},\hat{s}_{t-1})~\tikz[baseline]{\node[draw=black,fill=white,text=black,circle,inner sep=1pt,scale=0.7] {2};}$\end{tabular} \\ \cline{3-6} 
\multicolumn{1}{c|}{} &
  \multicolumn{1}{c|}{} &
  \multicolumn{1}{c|}{\multirow{2}{*}{Stochastic State}} &
  Train &
  $s_{t} = q_\theta(h_{t}, e_{t})$ &
  $\hat{s}_{t} = \hat{q}_\theta(\hat{h}_{t}, \hat{e}_{t})$ \\ \cline{4-6} 
\multicolumn{1}{c|}{} &
  \multicolumn{1}{c|}{} &
  \multicolumn{1}{c|}{} &
  Infer &
  $s_{t} = p_\theta(h_{t})$ &
  $\hat{s}_{t} = \hat{p}_\theta(\hat{h}_{t})$ \\ \cline{2-6}
\multicolumn{1}{c|}{} &
  \multicolumn{1}{c|}{\multirow{3}{*}{Heads}} &
  \multicolumn{2}{c|}{Reward} &
  $r_{t}=\text{Reward}(h_{t}, s_{t})$ &
  / \\ \cline{3-6} 
\multicolumn{1}{c|}{} &
  \multicolumn{1}{c|}{} &
  \multicolumn{2}{c|}{Decoder} &
  $d_{t}=\text{Decoder}(h_{t}, s_{t})$ &
  $\hat{d}_{t}=\hat{\text{Decoder}}(\hat{h}_{t}, \hat{s}_{t})$ \\ \cline{3-6} 
\multicolumn{1}{c|}{} &
  \multicolumn{1}{c|}{} &
  \multicolumn{2}{c|}{Continue} &
  $c_{t}=\text{Continue}(h_{t}, s_{t})$ &
  / \\ \bottomrule
\end{tabular}
}
\end{table*}

Instead of directly adopting a classic MBRL structure like Dreamer V3~\cite{hafner2023mastering}, \textbf{\textit{Raw2Drive}} is a dual-stream MBRL framework, as shown in Figure~\ref{fig:method}. It consists of four key components: two world models and two corresponding policy models. Detailed symbol definitions are in Table~\ref{table:notation}. 

The rationale behind the two stream design is that: \textbf{raw sensor data (e.g., high resolution multi-view videos) are high-dimensional and complex, presenting significant challenges to directly train a world model with low error}~\cite{brooks2024video, xia2024rgbd}. Thus, rather than building the raw sensor world model from scratch, we first build the privileged stream as an auxiliary since the learning of the world model under structured and low-dimensional conditions is much easier and there are already some successes~\cite {li2024think}. The raw sensor stream's learning process is guided by the trained privileged stream, which eases the task of extracting the decision-making related information from raw sensor data. The \textbf{\textit{Guidance Mechanism}} is carefully designed to avoid cumulative error and train-val gaps. 

\subsection{Privileged Stream}
\textbf{Privileged Input}: As shown in Figure~\ref{fig:method} (a), similar to Roach~\cite{zhang2021end} and Think2Drive~\cite{li2024think}, we utilize time-sequenced BEV semantic masks $o_t$ as input. 
 Since it is commonly used in existing works, we leave details of the privileged input in Appendix~\ref{sec:dual_stream}.

\begin{figure}[!]
    \centering
    \includegraphics[width=1\linewidth]{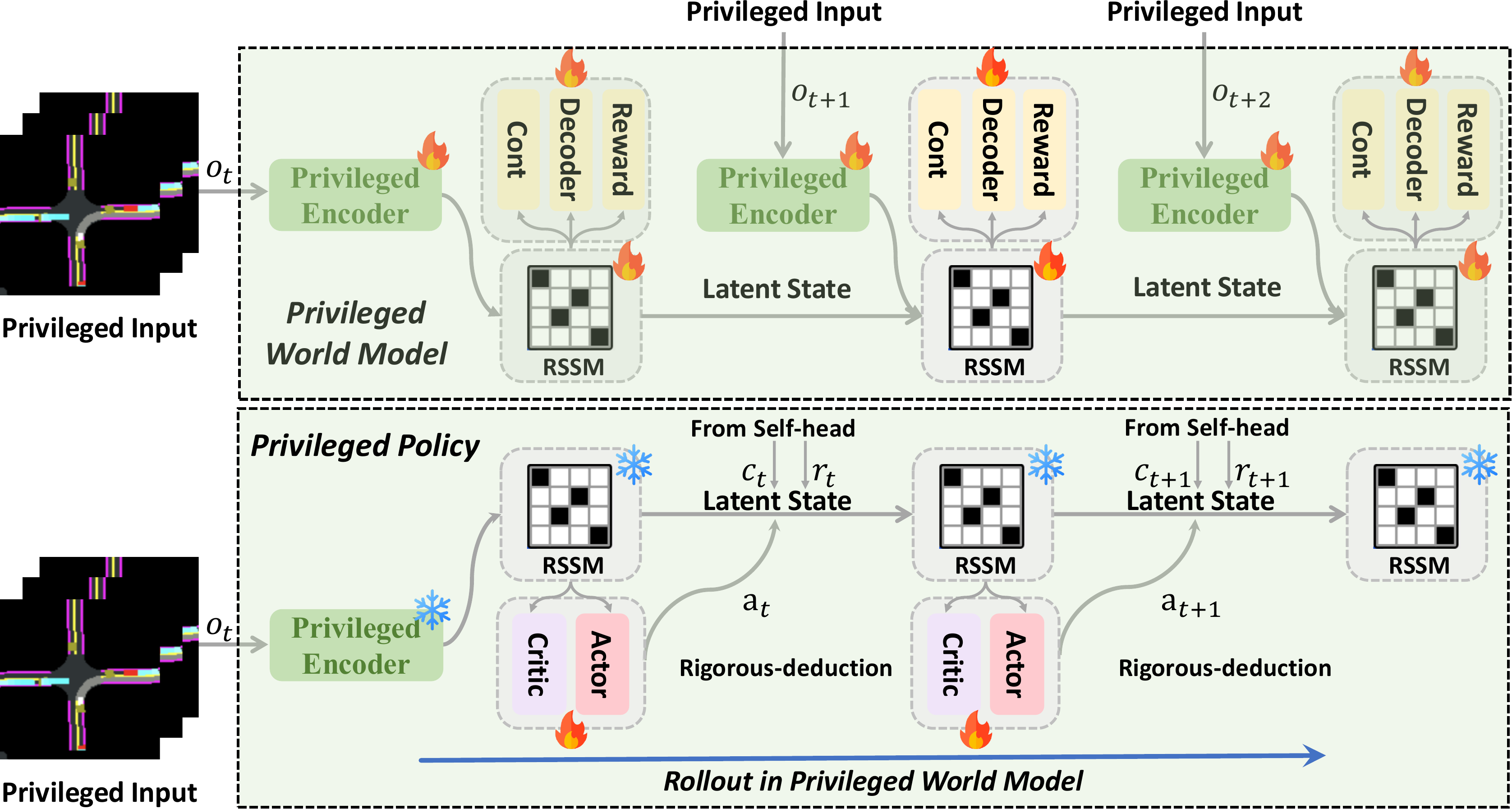}
    \caption{\textbf{Training of Privileged World Model  and Policy}. The privileged world model $\text{WM}$ is trained with time-sequenced BEV semantic masks as inputs. The privileged policy $\pi$ is trained through rollouts in the privileged world model. The reward $r_t$ and continuous flag $c_t$ are generated by the heads of the privileged world model with the privileged input $o_t$.}
    \label{fig:p_wm_learning}
\end{figure}

\noindent\textbf{Privileged World Model}: 
As shown in the upper part of Figure~\ref{fig:p_wm_learning}, the same as Dreamer V3~\cite{hafner2023mastering}, the privileged world model $\text{WM}$ is composed of the Encoder, Recurrent State-Space Model (RSSM)~\cite{doerr2018probabilistic} and three heads, defined in the left part of Table~\ref{table:notation}. These components are used (I) for rollout to train the privileged policy by RL; (II) to guide the training of the end-to-end raw sensor stream.

\noindent\textbf{Privileged  Policy}: 
As shown in the upper part of Figure~\ref{fig:p_wm_learning}, the privileged policy is composed of actor-critic networks and is trained through rollouts, same as in Dreamer V3~\cite{hafner2020dreamer}.

\subsection{Raw Sensor Stream}
\label{sec:raw_sensor_stream}
\noindent\textbf{Raw Sensor Input}: As in Figure~\ref{fig:method} (a), raw sensor input $\hat{o}_t$ in this work is multi-view images and IMU. We adopt BEVFormer~\cite{li2022bevformer} as the encoder $\hat{\text{Enc}}(\hat{o}_t)$ of the raw sensor stream, which outputs grid-shaped BEV features for the ease of receiving guidance from privileged streams. The details of the inputs are in Appendix~\ref{sec:dual_stream}.

\begin{figure}[!]
    \centering
    \includegraphics[width=1\linewidth]{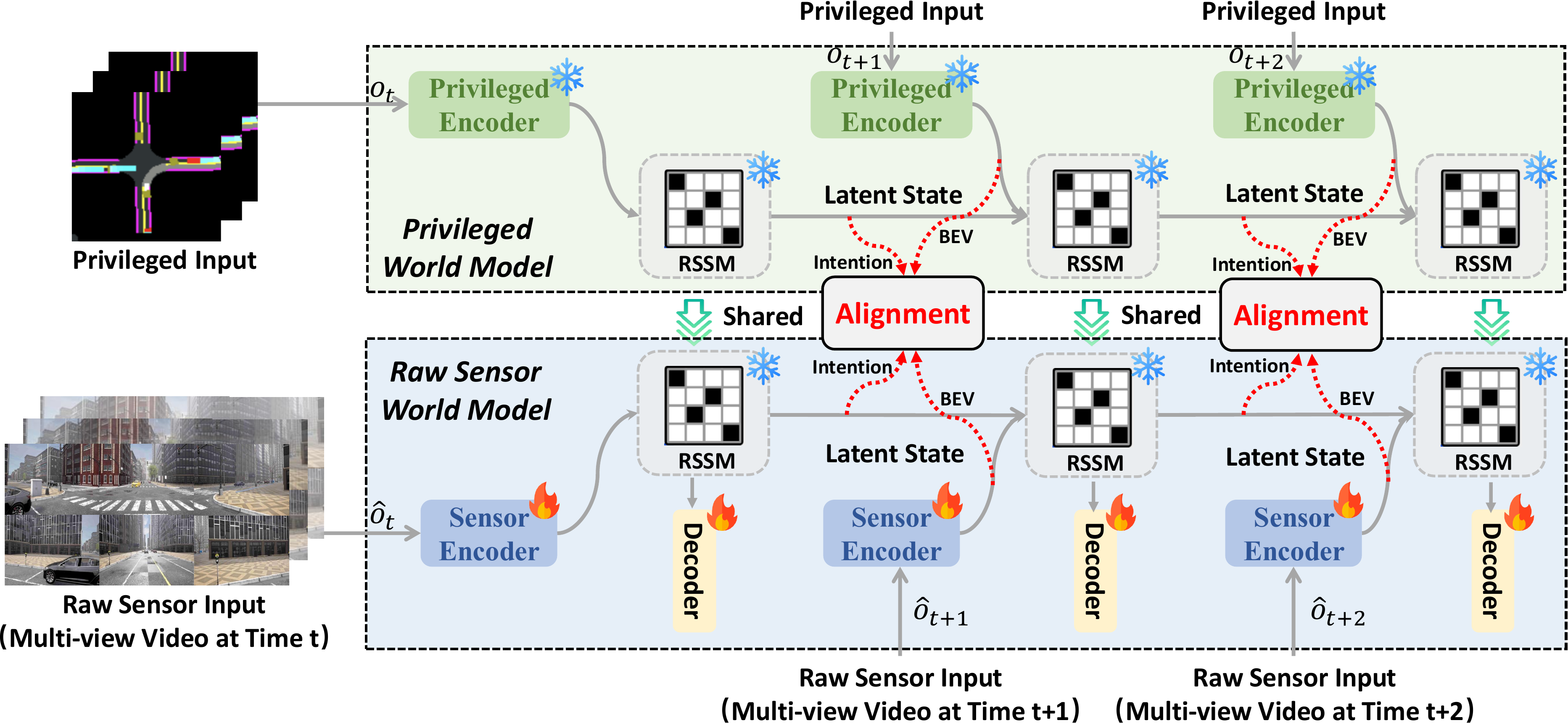}
    \caption{\textbf{Training  of Raw Sensor World Model}. During training, the spatial-temporal feature of the privileged world model serves as supervision instead of reconstructing multi-view video so that the learning could focus on decision related information. 
    RSSM parameters are initialized from the privileged world model. }
    \label{fig:r_wm_learning}
\end{figure}

\noindent\textbf{Raw Sensor World Model}:
As shown in Figure~\ref{fig:r_wm_learning}, the raw sensor world model has a similar architecture as the privileged world model except for the encoder $\hat{Enc}$ and heads. The different encoders are used to process different inputs. As for the heads, \textbf{we only use the decoder head which provides supervision signals based on BEV semantic masks} instead of directly reconstructing multi-view videos. Additionally, we find that \textbf{learning rewards or continuous flag (both only one scalar) could be harmful for the raw sensor stream}. Specifically, we observe that two adjacent similar frames could have very different reward and continuous labels, which is confusing and thus hampers convergence. We analyze these effects in Section~\ref{sec:raw_sensor_wm_heads}. 

\begin{figure}[!t]
    \centering
    \includegraphics[width=1\linewidth]{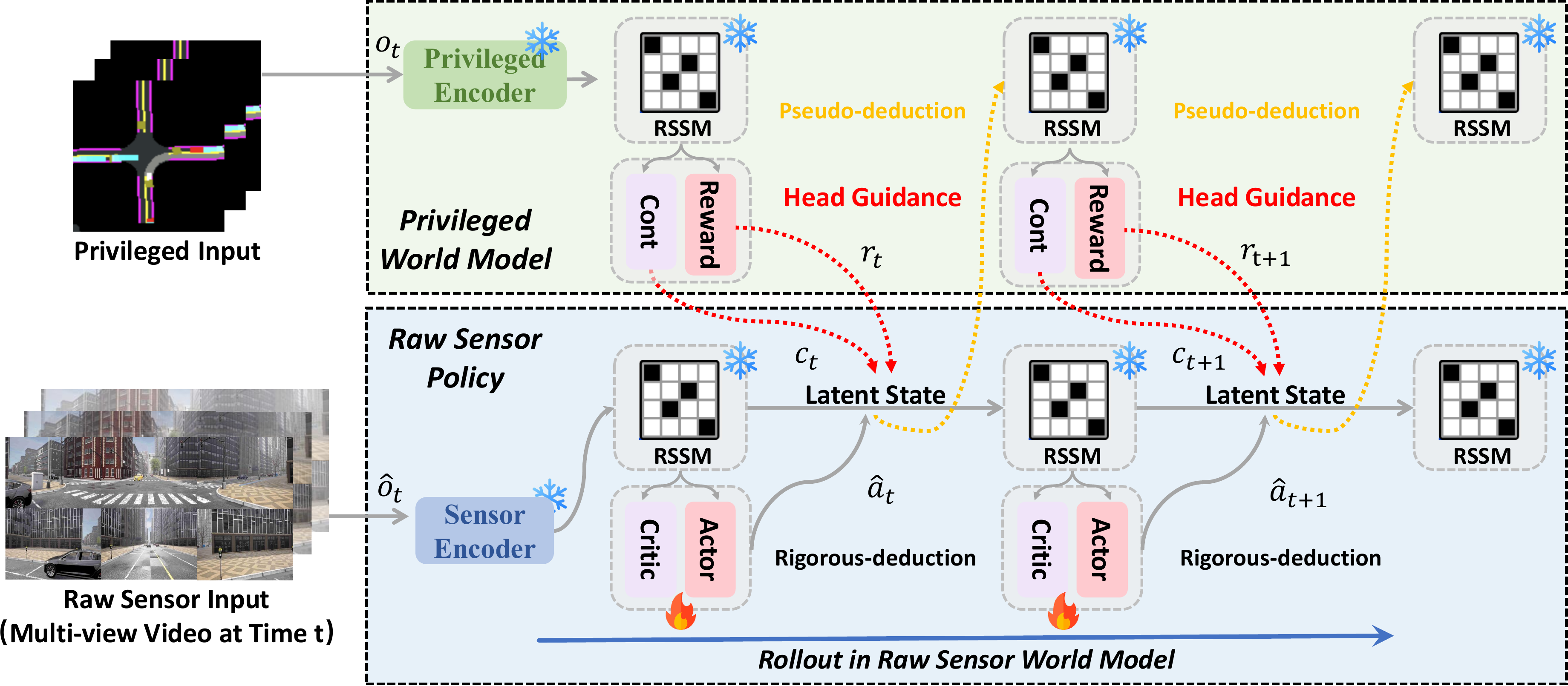}
    \caption{\textbf{Training of Raw Sensor Policy.} The raw sensor policy is trained through RL within the dual-stream world model. The raw model operates under strict deduction, while the reward $r_t$ and continuation flag $c_t$ are derived from the privileged model via pseudo-deduction.
    }
    \label{fig:r_bp_learning}
\end{figure}

\noindent\textbf{Raw Sensor Policy}:
As in Figure~\ref{fig:r_bp_learning}, the raw sensor policy is trained by RL with the dual-stream world model. During rollouts, the guidance mechanism detailed in Section~\ref{sec:guidance_mechanism} ensures consistency between the two world models in future predictions. We adopt the heads in the privileged world model to obtain the reward and continuous flag to provide more accurate and stable supervisor signals.

\subsection{Guidance Mechanism}
\label{sec:guidance_mechanism}
In the previous section, we introduced Raw2Drive's dual-stream MBRL framework. In this section, we give details of the guidance mechanism about (I) how the privileged stream guide the learning process of both the raw sensor world and policy; (II) how to alleviate cumulative errors and train-val gaps of the dual-stream paradigm.

\begin{figure}[htbp]
\centering
\includegraphics[width=1\textwidth]{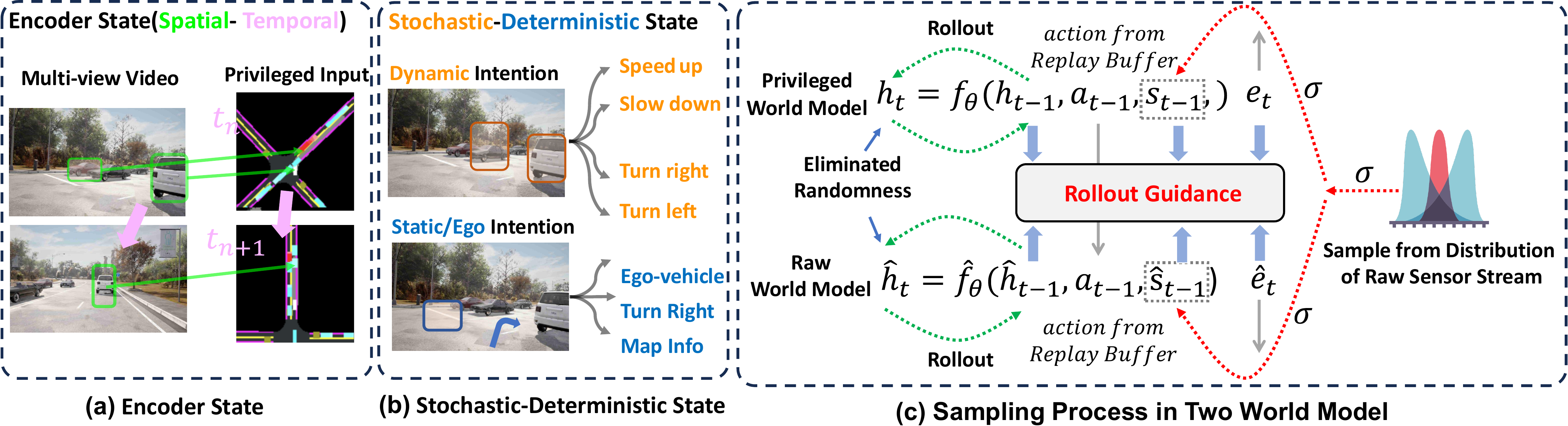}
\caption{\textbf{State Variables Aligned and Sampling Process in the Rollout Guidance}. (a) The encoder state is aligned temporally and spatially. (b) The deterministic state and stochastic state is aligned to maintain dynamic and static intention consistency. (c) \textbf{Eliminating Cumulative Errors Caused By Sampling}. During the rollout process, when deducting next states, we only sample once for the stochastic state - from the distribution of the raw sensor stream. The sampled state is fed into both streams to eliminate the randomness and thus the alignment is more stable.}
\label{fig:rollout}    
\end{figure}
\textbf{Rollout Guidance.}\label{sec:rollout_guidance}\quad
In MBRL framework, the rollout of world model is crucial for effective policy learning. Thus, under the proposed dual-stream framework, the first type of guidance is Rollout Guidance, which adopts the trained privileged world model provides supervisions during the entire rollout process of the raw sensor world model. Specifically, as shown in Figure~\ref{fig:rollout}, there are three components in the rollout process: (1) Encoded State $\hat{e}$, (2) Deterministic State $\hat{h}$, (3) Stochastic State $\hat{s}$.
At each timestep $t$, both stream encode the observation (either privileged or raw sensor) into Encoded State.  However, due to the high dimensionality and inherent redundancy of multi-view video data, its encoded state $\hat{e}$ of the raw sensor stream may exhibit instability and errors. To address this, we introduce a loss function to align it with the privileged encoded state $e$, ensuring spatial consistency at each timestep, called \textbf{Spatial-Temporal Alignment Loss}.  

Meanwhile, as illustrated in the upper part of Figure \ref{fig:method} (c), and Figure~\ref{fig:rollout} (b), during the rollout process, the deterministic and stochastic states serve distinct modeling purposes. The deterministic state primarily predicts the ego vehicle’s state, while the stochastic state focuses on anticipating the behaviors of other traffic participants, such as sudden braking or deceleration. To ensure the consistency of states in the rollout of the two world models, additional supervisory signals are introduced to ensure alignment with the privileged world model. Specifically, the deterministic state employs L2 loss to maintain prediction consistency, whereas the stochastic state is sampled from an independent one-hot distribution. KL divergence~\cite{kullback1951information} is then used to constrain the distributions of the two information streams, ensuring they remain as similar as possible. Furthermore, alignment is enforced across all timesteps, called \textbf{Abstract-State Alignment Loss}. The overall loss function is:
\begin{align}
\mathcal{L}_{\text{Rollout}} = & \beta_{e}\sum_t \sum_{i=0}^{\text{grid num}} \text{MSE}(e_t^{i}, \hat{e}_t^{i}) + \sum_t(\beta_{s} \text{KL}(s_t, \hat{s}_t) + \beta_{h} \text{MSE}(h_t, \hat{h}_t))
\end{align}
where grid num means the number of grids under BEV-view and $\beta_{e}$, $\beta_{h}$, $\beta_{s}$ are the loss weights of encoder state, deterministic state, and stochastic state. Notably, in the standard RSSM~\cite{hafner2023mastering}, the deduction of stochastic states employs sampling during rollout. \textbf{This sampling process can be detrimental for the training of raw sensor world model, as difference caused by randomness during sampling between the two models could accumulate over time}. As a result, in the later timesteps,  using states from the privileged world model as supervision signals is confusing for the raw sensor world model to align with. To this end, as shown in Figure~\ref{fig:rollout}, we only conduct sampling from the distribution of raw sensor stream and directly feed the sampled variable into the privileged world model to deduct its next state. In this way, the cumulative errors caused by randomness is eliminated and thus is beneficial for the trianing of raw sensor model.

\textbf{Head Guidance.}\quad
\label{sec:head_guidance}
During the training of the raw sensor world model, as mentioned in Sec~\ref{sec:raw_sensor_stream}, only the decoder head is trained. We omit the reward head and continuous head learning because directly training the two heads with raw sensor inputs has a convergence issue. Adjacent frames in the video are highly similar while value of reward and continuous flag could fluctuate abruptly, as in Figure~\ref{fig:reward_cont}. As a result, it is difficult for the network to learn stable patterns.

For MBRL, reward and continuous flag play a crucial role in guiding raw policy training. Thus, \textbf{we use the accurate reward and continuous flag from the privileged world model during the training of raw sensor policy.} As shown in Figure~\ref{fig:method} (c) Lower Part and Figure~\ref{fig:r_bp_learning} , at each timestep $t$, the raw sensor world model executes the action $a_t$ obtained from the raw policy $\hat{\pi}$, transitioning the system to next latent state.  At the same time, the privileged world model conducts the same action to rollout. Since \textbf{\textit{Rollout Guidance}} imposes the consistency between the two world models, the reward $r_t$ and the continuation flag $c_t$ from the privileged world model could be directly used. Finally, the resulting sequence serves as the training data for optimizing the raw sensor policy. Notably, we also adopt the technique to eliminate randomness mentioned in Section~\ref{sec:head_guidance} so that the reward and continuous flag is accurate. Furthermore, we also use the trained privileged policy to collect to replay buffer and to distill action distributions to raw sensor policy. We summarize our training pipeline in Appendix~\ref{sec:train_pipeline}.

\section{Experiments}
\label{sec:experiments}
\subsection{Datasets and Benchmark}
We employ the CARLA simulator~\cite{dosovitskiy2017carla} (version 0.9.15.1) for closed-loop driving performance evaluation. Note that during evaluation, the model only has access to raw sensor observations and is prohibited from utilizing privileged observations. Experimental details are in Appendix~\ref{sec:exp}.

\textbf{Leaderboard 2.0~\cite{dosovitskiy2017carla}:} It includes two long routes, \textit{validation} and \textit{devtest}. Each of which comprises several routes with lengths of 7-10 kilometers and containing a series of complex corner cases. \textit{As driving inherently follows a Markov decision process~\cite{jaeger2025carl}, evaluating performance over long routes is unnecessary. Moreover, the penalty mechanism employed in scoring~\cite{li2024think} fails to accurately reflect the true evaluation capability of the model}.

\textbf{Bench2Drive~\cite{jia2024bench}:} A more comprehensive and fair benchmark, it includes 220 short routes with one challenging corner case per route for analysis of different driving abilities. Following Bench2Drive, we use 1,000 routes under diverse weather conditions for RL training. 

\begin{table}[]
\centering
\caption{\textbf{Performance on Carla Official Town13 Validation and Devtest Benchmark}. *denotes expert feature distillation. As discussed in carla-garage~\cite{Zimmerlin2024ArXiv} and Section~\ref{sec:sota}, long routes evaluation in Leaderboard 2.0 can't reflect the actual driving performance~\cite{jia2024bench}. \label{tab:devtest_validation}}
 \resizebox{\textwidth}{!}{
\begin{tabular}{c|c|c|c|cccccc}
\toprule
\multirow{3}{*}{\textbf{Method}} &
\multirow{3}{*}{\textbf{Venue}} &
\multirow{3}{*}{\textbf{Scheme}} &
\multirow{3}{*}{\textbf{Modality}} &
  \multicolumn{6}{c}{\textbf{Closed-loop Metric}} \\ \cline{5-10} 
 & & & &
  \multicolumn{2}{c|}{DS $\uparrow$} &
  \multicolumn{2}{c|}{RC (\%) $\uparrow$} &
  \multicolumn{2}{c}{IS $\uparrow$} \\ \cline{5-10} 
 & & & &
  \multicolumn{1}{c|}{Validation} &
  \multicolumn{1}{c|}{Devtest} &
  \multicolumn{1}{c|}{Validation} &
  \multicolumn{1}{c|}{Devtest} &
  \multicolumn{1}{c|}{Validation} &
  Devtest \\ \hline
AD-MLP~\cite{zhai2023rethinking} &
  Arxiv 2023 &
  IL &
  State &
  \multicolumn{1}{c|}{0.00} &
  \multicolumn{1}{c|}{0.00} &
  \multicolumn{1}{c|}{0.00} &
  \multicolumn{1}{c|}{0.00} &
  \multicolumn{1}{c|}{0.00} &
  0.00 \\
UniAD-Base~\cite{hu2023planning} &
  CVPR 2023 &
  IL &
  Image &
  \multicolumn{1}{c|}{0.15} &
  \multicolumn{1}{c|}{0.00} &
  \multicolumn{1}{c|}{0.51} &
  \multicolumn{1}{c|}{0.07} &
  \multicolumn{1}{c|}{0.23} &
  0.04 \\
VAD~\cite{jiang2023vad} &
  ICCV 2023 &
  IL &
  Image &
  \multicolumn{1}{c|}{0.17} &
  \multicolumn{1}{c|}{0.00} &
  \multicolumn{1}{c|}{0.49} &
  \multicolumn{1}{c|}{0.06} &
  \multicolumn{1}{c|}{0.31} &
  0.04 \\ 
DriveTrans~\cite{jia2025drivetransformer} &
  ICLR 2025 &
  IL &
  Image &
  \multicolumn{1}{c|}{0.85} &
  \multicolumn{1}{c|}{0.68} &
  \multicolumn{1}{c|}{1.42} &
  \multicolumn{1}{c|}{2.13} &
  \multicolumn{1}{c|}{0.33} &
  0.35 \\
TCP-traj*~\cite{wu2022trajectoryguided} &
  NeurIPS 2022 &
  IL &
  Image &
  \multicolumn{1}{c|}{0.31} &
  \multicolumn{1}{c|}{0.02} &
  \multicolumn{1}{c|}{0.89} &
  \multicolumn{1}{c|}{0.11} &
  \multicolumn{1}{c|}{0.24} &
  0.05 \\
ThinkTwice*~\cite{jia2023thinktwice} &
  CVPR 2023 &
  IL &
  Image &
  \multicolumn{1}{c|}{0.50} &
  \multicolumn{1}{c|}{0.64} &
  \multicolumn{1}{c|}{1.23} &
  \multicolumn{1}{c|}{1.78} &
  \multicolumn{1}{c|}{0.35} &
  \textbf{0.43} \\
DriveAdapter*~\cite{jia2023driveadapter} &
  ICCV 2023 &
  IL &
  Image &
  \multicolumn{1}{c|}{0.92} &
  \multicolumn{1}{c|}{0.87} &
  \multicolumn{1}{c|}{1.52} &
  \multicolumn{1}{c|}{2.43} &
  \multicolumn{1}{c|}{0.42} &
  0.37 \\ \hline
  \rowcolor{gray!20} Raw2Drive (Ours) &
  - &
  RL &
  Image &
  \multicolumn{1}{c|}{\textbf{4.12}} &
  \multicolumn{1}{c|}{\textbf{3.56}} &
  \multicolumn{1}{c|}{\textbf{9.32}} &
  \multicolumn{1}{c|}{\textbf{6.04}} &
  \multicolumn{1}{c|}{\textbf{0.43}} &
  0.42 \\
  \textcolor{gray!50}{Think2Drive}~\cite{li2024think} &
  \textcolor{gray!50}{ECCV 2024} &
  \textcolor{gray!50}{RL Expert} &
  \textcolor{gray!50}{Privileged*} &
  \multicolumn{1}{c|}{\textcolor{gray!50}{43.8}} &
  \multicolumn{1}{c|}{\textcolor{gray!50}{56.8}} &
  \multicolumn{1}{c|}{\textcolor{gray!50}{78.2}} &
  \multicolumn{1}{c|}{\textcolor{gray!50}{98.6}} &
  \multicolumn{1}{c|}{\textcolor{gray!50}{0.73}} &
  \textcolor{gray!50}{0.92} \\ \bottomrule
\end{tabular}
}
\end{table}

\begin{table*}[]
\centering
\caption{\textbf{Performance on Bench2Drive Multi-Ability Benchmark}. * denotes expert feature distillation. IL represents imitation learning. RL represents reinforcement learning. RL expert Think2Drive~\cite{li2024think} uses privileged information for training. \label{tab:ability}}
 \resizebox{\textwidth}{!}{
\begin{tabular}{l|c|c|c|cccccc}
\toprule
\multirow{2}{*}{\textbf{Method}} &
\multirow{2}{*}{\textbf{Venue}} &
\multirow{2}{*}{\textbf{Scheme}} &
\multirow{2}{*}{\textbf{Modality}} &
\multicolumn{6}{c}{\textbf{Ability} (\%) $\uparrow$} \\ \cline{5-10}
 & & & &
  \multicolumn{1}{c|}{Merging} &
  \multicolumn{1}{c|}{Overtaking} &
  \multicolumn{1}{c|}{Emergency Brake} &
  \multicolumn{1}{c|}{Give Way} &
  \multicolumn{1}{c|}{Traffic Sign} &
  \textbf{Mean} \\ \hline
TCP-traj*~\cite{wu2022trajectoryguided} &
  NeurIPS 2022 &
  IL &
  Image &
  \multicolumn{1}{c|}{8.89} &
  \multicolumn{1}{c|}{24.29} &
  \multicolumn{1}{c|}{51.67} &
  \multicolumn{1}{c|}{40.00} &
  \multicolumn{1}{c|}{46.28} &
  34.22 \\
AD-MLP~\cite{zhai2023rethinking} &
  Arxiv 2023 &
  IL &
  State &
  \multicolumn{1}{c|}{0.00} &
  \multicolumn{1}{c|}{0.00} &
  \multicolumn{1}{c|}{0.00} &
  \multicolumn{1}{c|}{0.00} &
  \multicolumn{1}{c|}{4.35} &
  0.87 \\
UniAD-Base~\cite{hu2023planning} &
  CVPR 2023 &
  IL &
  Image &
  \multicolumn{1}{c|}{14.10} &
  \multicolumn{1}{c|}{17.78} &
  \multicolumn{1}{c|}{21.67} &
  \multicolumn{1}{c|}{10.00} &
  \multicolumn{1}{c|}{14.21} &
  15.55 \\
ThinkTwice*~\cite{jia2023thinktwice} &
  CVPR 2023 &
  IL &
  Image &
  \multicolumn{1}{c|}{27.38} &
  \multicolumn{1}{c|}{18.42} &
  \multicolumn{1}{c|}{35.82} &
  \multicolumn{1}{c|}{\textbf{50.00}} &
  \multicolumn{1}{c|}{54.23} &
  37.17 \\
VAD~\cite{jiang2023vad} &
  ICCV 2023 &
  IL &
  Image &
  \multicolumn{1}{c|}{8.11} &
  \multicolumn{1}{c|}{24.44} &
  \multicolumn{1}{c|}{18.64} &
  \multicolumn{1}{c|}{20.00} &
  \multicolumn{1}{c|}{19.15} &
  18.07 \\
DriveAdapter*~\cite{jia2023driveadapter} &
  ICCV 2023 &
  IL &
  Image &
  \multicolumn{1}{c|}{28.82} &
  \multicolumn{1}{c|}{26.38} &
  \multicolumn{1}{c|}{48.76} &
  \multicolumn{1}{c|}{\textbf{50.00}} &
  \multicolumn{1}{c|}{56.43} &
  42.08 \\ 
DriveTrans~\cite{jia2025drivetransformer} &
  ICLR 2025 &
  IL &
  Image &
  \multicolumn{1}{c|}{17.57} &
  \multicolumn{1}{c|}{35.00} &
  \multicolumn{1}{c|}{48.36} &
  \multicolumn{1}{c|}{40.00} &
  \multicolumn{1}{c|}{52.10} &
  38.60 \\
\hline
\rowcolor{gray!20} Raw2Drive (Ours) &
  - &
  RL &
  Image &
  \multicolumn{1}{c|}{\textbf{43.35}} &
  \multicolumn{1}{c|}{\textbf{51.11}} &
  \multicolumn{1}{c|}{\textbf{60.00}} &
  \multicolumn{1}{c|}{\textbf{50.00}} &
  \multicolumn{1}{c|}{\textbf{62.26}} &
  \textbf{53.34} \\
\textcolor{gray!50}{Think2Drive}~\cite{li2024think} &
  \textcolor{gray!50}{ECCV 2024} &
  \textcolor{gray!50}{RL Expert} &
  \textcolor{gray!50}{Privileged*} &
  \multicolumn{1}{c|}{\textcolor{gray!50}{81.27}} &
  \multicolumn{1}{c|}{\textcolor{gray!50}{83.92}} &
  \multicolumn{1}{c|}{\textcolor{gray!50}{90.24}} &
  \multicolumn{1}{c|}{\textcolor{gray!50}{90.00}} &
  \multicolumn{1}{c|}{\textcolor{gray!50}{87.67}} &
  \textcolor{gray!50}{86.26} \\ \bottomrule
  
\end{tabular}
}
\end{table*}

\subsection{Metric}
We employ the official metrics of CARLA~\cite{dosovitskiy2017carla} for evaluation. \textbf{Infraction Score} (IS) measures the number of infractions made along the route, accounting for interactions with pedestrians, vehicles, road layouts, red lights, etc. \textbf{Route Completion} (RC) indicates the percentage of the route completed by the autonomous agent. \textbf{Driving Score (DS)}, is calculated as the product of Route Completion and Infraction Score. In Bench2Drive~\cite{jia2024bench}, additional metric \textbf{Success Rate} is used, which measures the proportion of successfully completed routes within the allotted time and without traffic violations.

\begin{table}[tb!]
\centering
\caption{\textbf{Results on Bench2Drive Closed-loop Benchmark.} *denotes expert feature distillation. * denotes expert feature distillation. IL represents imitation learning. RL represents reinforcement learning. RL expert Think2Drive~\cite{li2024think} uses privileged information for training.
\label{tab:perfomance_b2d}}
\resizebox{\linewidth}{!}{
\begin{tabular}{l|c|c|c|cccc}
\toprule
\multirow{2}{*}{\textbf{Method}} & \multirow{2}{*}{Venue} & \multirow{2}{*}{Scheme} & \multirow{2}{*}{Modality} & \multicolumn{4}{c}{\textbf{Closed-loop Metric}} \\ \cline{5-8} 
                                                         & & & & \makecell{DS $\uparrow$} & \makecell{SR(\%) $\uparrow$}  & \makecell{Efficiency $\uparrow$} & \makecell{Comfort $\uparrow$} \\ \hline
TCP-traj*~\cite{wu2022trajectoryguided}  & NeurIPS 2022 & IL & Image & 59.90 & 30.00 & 76.54          & 18.08          \\
AD-MLP~\cite{zhai2023rethinking}        & Arxiv 2023 & IL & State &18.05 & 0.00  & 48.45  & 22.63          \\
UniAD-Base~\cite{hu2023planning}         & CVPR 2023 & IL & Image &45.81 & 16.36 & 129.21 & 43.58          \\
VAD~\cite{jiang2023vad}                  & ICCV 2023 & IL & Image &42.35 & 15.00 & 157.94 & \textbf{46.01} \\ 
ThinkTwice*~\cite{jia2023thinktwice}     & CVPR 2023& IL & Image & 62.44 & 31.23 & 69.33          & 16.22          \\
DriveAdapter*~\cite{jia2023driveadapter} & ICCV 2023 & IL & Image & 64.22           & 33.08              & 70.22                 & 16.01                  \\ 
GenAD~\cite{zheng2024genad}              & ECCV 2024 & IL & Image & 44.81  & 15.90  & - & - \\
DriveTrans~\cite{jia2025drivetransformer}  & ICLR 2025 & IL & Image &63.46 & 35.01 & 100.64 & 20.78  \\
MomAD~\cite{song2025don}                 & CVPR 2025 & IL & Image & 44.54  & 16.71  & 170.21 & 48.63                  \\
\hline
\rowcolor{gray!20} Raw2Drive (Ours)    & - & RL & Image & \textbf{71.36}           & \textbf{50.24}              & \textbf{214.17}                 & 22.42 \\                  
\textcolor{gray!50}{Think2Drive}~\cite{li2024think} & \textcolor{gray!50}{ECCV 2024} & \textcolor{gray!50}{RL Expert} & \textcolor{gray!50}{Privileged} & \textcolor{gray!50}{91.85}           & \textcolor{gray!50}{85.41}              & \textcolor{gray!50}{269.14}                 &  \textcolor{gray!50}{25.97}       \\ \bottomrule
\end{tabular}
}
\end{table}

\subsection{Comparison with State-of-the-Art Works}
\label{sec:sota}
As in Table~\ref{tab:devtest_validation}, similar to the conclusions of Bench2Drive, the long-road evaluation in CARLA 2.0 fails to accurately reflect real driving performance due to its cumulative penalty scoring mechanism. In its devtest/validation routes, the first scenario is \textbf{\textit{ParkingExit}}, where traditional imitation learning methods struggle to solve the task. This is primarily because these models typically rely on L2/L1 loss for training while \textbf{\textit{ParkingExit}} requires large steering angles, which are challenging to learn effectively due to the inherent imbalance in the data distribution. Additionally, when a collision causes a blockage, the evaluation stops immediately, preventing the continuation of subsequent scenarios. This does not mean that the performance is poor in other scenarios. Therefore, we mainly focus on the short-route closed-loop evaluation results of Bench2Drive. As shown in Table~\ref{tab:ability}, our method achieves SOTA performance in multi-ability benchmark. In Bench2Drive's closed-loop benchmark, as shown in Table~\ref{tab:perfomance_b2d}, Raw2Drive achieves SOTA performance in raw sensor end-to-end methods.

\subsection{Ablation Study}
\label{sec:ablation}
Following Bench2Drive~\cite{jia2024bench, githubGitHubThinklabSJTUBench2Drive} and DriveTransformer~\cite{jia2025drivetransformer}, the tiny validation set \textbf{\textit{Dev10}} comprises 10 carefully selected clips from the official 220 routes. These clips are chosen to be both challenging and representative, with low variance. 
\textit{To avoid overfitting Bench2Drive-220, we use Dev10 for ablations, repeating each experiment 3 times and reporting the average}.

\textbf{Choice of Heads in Raw Sensor World Model.}\quad
\label{sec:raw_sensor_wm_heads}
Due to the high-dimensional redundancy of image information and the fact that both the reward and continue flags are represented as scalars, the network struggles to achieve stable convergence during training. To further investigate this, we conduct an ablation study on the raw sensor world model heads. As shown in Table~\ref{tab:wm_head}, the decoder head provides effective supervision, significantly enhancing the learning of the raw sensor world model. In contrast, the reward and continuation heads introduce ambiguity that hinders convergence, leading to suboptimal world model training and degraded policy performance. We provide a detailed visualization in the Appendix~\ref{sup:perception_confuse}.

\begin{table}[H]
    \centering
    \begin{minipage}[t]{0.46\textwidth} 
    \caption{\textbf{Ablation on the Raw Sensor World Model Heads}. Only the decoder head is used.
    }
    \centering
    \resizebox{\textwidth}{!}{
    \begin{tabular}{ccc|cc}
    \toprule
    \multicolumn{3}{c|}{\textbf{Heads}}                                   & \multirow{2}{*}{\textbf{DS$\uparrow$}} & \multirow{2}{*}{\textbf{SR$\uparrow$}} \\ \cline{1-3}
    \multicolumn{1}{c|}{Decoder} & \multicolumn{1}{c|}{Reward} & Continue &                                        &                                        \\ \hline
    \multicolumn{1}{c|}{$\times$} & \multicolumn{1}{c|}{$\times$} &  $\times$ &  17.4 & 1.2/10  \\ \hline
    \rowcolor{gray!20} \multicolumn{1}{c|}{\checkmark} & \multicolumn{1}{c|}{$\times$} &  $\times$ &  \textbf{83.5} & \textbf{7.5/10}  \\ \hline
    \multicolumn{1}{c|}{\checkmark} & \multicolumn{1}{c|}{$\checkmark$} &  $\times$ & 46.6 & 3.4/10  \\ \hline
    \multicolumn{1}{c|}{\checkmark} & \multicolumn{1}{c|}{$\checkmark$} & $\checkmark$ &  34.5 & 2.2/10  \\ \bottomrule
    \end{tabular}
    }
    \label{tab:wm_head}
    \end{minipage} 
    \hfill 
     \begin{minipage}[t]{0.50\textwidth}
     \centering
     \caption{\textbf{Ablation on the Abstract-State Alignment in Rollout Guidance}.}
     \resizebox{\textwidth}{!}{
    \begin{tabular}{ccc|cc}
    \toprule
    \multicolumn{3}{c|}{\textbf{Latent State}}                                   & \multirow{2}{*}{\textbf{DS$\uparrow$}} & \multirow{2}{*}{\textbf{SR$\uparrow$}} \\ \cline{1-3}
    \multicolumn{1}{c|}{Encoder} & \multicolumn{1}{c|}{Deterministic} & Stochastic &                                        &                                        \\ \hline
    \multicolumn{1}{c|}{$\times$} & \multicolumn{1}{c|}{$\times$} &  $\times$ & 0.0 & 0.0/10  \\ \hline
    \multicolumn{1}{c|}{$\checkmark$} & \multicolumn{1}{c|}{$\times$} &  $\times$ & 36.4 & 2.4/10  \\  \hline
    \multicolumn{1}{c|}{$\checkmark$} & \multicolumn{1}{c|}{$\checkmark$} &  $\times$ & 38.3 &  2.8/10 \\  \hline
    \rowcolor{gray!20} \multicolumn{1}{c|}{$\checkmark$} & \multicolumn{1}{c|}{$\checkmark$} & $\checkmark$ &  \textbf{83.5} & \textbf{7.5/10}  \\ \bottomrule
    \end{tabular}}
    \label{tab:rollout_g}
     \end{minipage} 
     \vspace{-10pt}
\end{table}

\textbf{Abstract-State Alignment in Rollout Guidance.}\label{sec:rollout_guidance}\quad
We conduct an ablation study to examine the role of Abstract-State Alignment in rollout guidance. As shown in Table~\ref{tab:rollout_g}, since the world model is trained during rollouts, any state misalignment can lead to discrepancies between the predictions of the two world models. Our results show that when any component of the rollout guidance is missing, the model can only handle simple tasks, such as moving straight or turning left, while failing to address more complex interactive corner cases. This misalignment significantly hinders the training of the raw sensor-based world model. All three components of rollout guidance components are thus essential for robust learning and performance. Loss comparisons are provided in Appendix~\ref{sup:rollout_guidance}. 

\textbf{Spatial-Temporal Alignment in Rollout Guidance.}\quad The results on Spatial-Temporal Alignment (based on Dev10) is shown in Table~\ref{tab:rollout_st}. Spatial Alignment ensures consistency between image representations and BEV representations—removing it is similar to “driving blind.” Temporal Alignment maintains the consistency of future predictions across time. Both components are essential and complementary for a stable world model rollout.

\begin{table}[h]
\centering
\caption{\textbf{Ablation on the Spatial-Temporal Alignment in Rollout Guidance.}}
\begin{tabular}{cc|c|c}
\toprule
\textbf{Spatial Alignment} & \textbf{Temporal Alignment} & \textbf{DS}$\uparrow$ & \textbf{SR(\%)}$\uparrow$ \\
\midrule
$\times$ & $\times$ & 0.0 & 0.0/10 \\
$\checkmark$ & $\times$ & 13.6 & 1.2/10 \\
$\times$ & $\checkmark$ & 9.24 & 0.8/10 \\
\rowcolor{gray!20} $\checkmark$ & $\checkmark$ & 83.5 & 7.5/10 \\
\bottomrule
\end{tabular}
\label{tab:rollout_st}
\end{table}

\textbf{Head Guidance.}\label{sec:head_guidance}\quad
For this ablation study, we assume that the raw sensor world model is trained with three heads. As shown in Table~\ref{tab:head_guidance}, both Setting I and Setting II follow this configuration, aligning with the setup in the last row of Table~\ref{tab:wm_head} and reflecting the same considerations and conclusions drawn in Ablation~\ref{sec:raw_sensor_wm_heads}. Our results indicate that head guidance provides some improvement in policy training. However, the additional heads increase training complexity and degrade performance. Thus, Raw2Drive adopts only the decoder head with head guidance.

\begin{table}[H]
    \centering
    \caption{
    \textbf{Ablation on head guidance for raw sensor policy learning.} D/R/C: decoder, reward, continuation; HG: head guidance.
    }
     \centering
     \resizebox{0.6\textwidth}{!}{
     \begin{tabular}{c|c|c|c|c|cc}
    \toprule
    \textbf{Method}    & \textbf{D} & \textbf{R} & \textbf{C} & \textbf{HG} & \textbf{DS$\uparrow$} & \textbf{SR$\uparrow$} \\ \hline
    \rowcolor{gray!20} Raw2Drive (Ours) & $\checkmark$           &   $\times$        &    $\times$         &      $\checkmark$          & \textbf{83.5} & \textbf{7.5/10}  \\ \hline
    Setting I & \checkmark           &   $\checkmark$    &    $\checkmark$     &      $\checkmark$            &   34.5 & 2.2/10    \\ \hline
    Setting II & \checkmark           &   $\checkmark$    &    $\checkmark$        &   $\times$               &  26.4 & 1.6/10   \\ 
    \bottomrule
    \end{tabular}}
    \label{tab:head_guidance}
    \vspace{-15pt}
\end{table}

\textbf{Shared Parameter.}\label{sec:para_shared}\quad
To investigate the impact of parameter sharing, we conduct an ablation study on whether the Recurrent State-Space Model (RSSM) and the decoder head should share parameters. As shown in Table~\ref{tab:para_shared}, we compare different configurations where these components are either shared or kept separate. Our findings indicate that sharing parameters between the RSSM and the decoder head leads to better performance. This suggests that parameter sharing facilitates more efficient representation learning, improving the consistency and generalization of the world model.

\begin{table}[H]
    \centering
    \begin{minipage}[t]{0.46\textwidth} 
     \caption{\textbf{Ablation Study on the Sharing Parameters (RSSM and Decoder Head)}.}
    \resizebox{0.9\linewidth}{!}{
    \begin{tabular}{c|cc}
    \toprule
    \textbf{Method} & \textbf{DS$\uparrow$ } & \textbf{SR$\uparrow$ } \\
    \hline
    \rowcolor{gray!20} Raw2Drive (Ours) & \textbf{83.5} & \textbf{7.5/10} \\ \hline
    w/o Shared RSSM    & 53.2 & 5.4/10 \\ \hline
    w/o Shared Head    & 65.6 & 6.1/10 \\
    \bottomrule
    \end{tabular}}
    \label{tab:para_shared}

    \end{minipage} 
    \hfill 
    \begin{minipage}[t]{0.50\textwidth}
    \caption{\textbf{Ablation Study on Raw Sensor Policy Training.} Compare two strategies: (1) directly using the privileged policy, and (2) fine-tuning the privileged policy on raw sensor inputs.}
    \resizebox{1\linewidth}{!}{ 
    \begin{tabular}{c|cc}
    \toprule
    \textbf{Method} & \textbf{DS$\uparrow$} & \textbf{SR$\uparrow$} \\
    \hline
    Directly Use Privileged Policy    & 58.4 & 5.6/10 \\ \hline
    \rowcolor{gray!20} Fine-tune Privileged Policy (Ours)  & \textbf{83.5} & \textbf{7.5/10} \\
    \bottomrule
    \end{tabular}}
    \label{tab:policy_finetune}
    \end{minipage} 
     \vspace{-15pt}
\end{table}

\textbf{Policy Finetuning.}\quad
To evaluate the necessity of the raw sensor world model, we conduct an ablation study by comparing policy fine-tuning with directly using the privileged policy. Specifically, we analyze the performance difference between policies trained with and without fine-tuning under the raw sensor world model. As shown in Table~\ref{tab:policy_finetune}, directly applying the pre-trained privileged policy without fine-tuning results in suboptimal performance, as the policy lacks adaptation to the raw sensor world model’s learned dynamics. 
In contrast, fine-tuning the policy within this world model leads to significant improvements, particularly in handling complex interactive tasks. This result underscores the importance of the raw sensor-based world model for effective policy learning and adaptation.

\textbf{Real-time Inference.} \quad We conducted latency analysis for each module. In end-to-end autonomous driving, the perception backbone (e.g., surround-view image encoder) typically dominates the overall latency. Our world model and policy are highly efficient (each under 2ms), while the raw sensor stream is mainly bottlenecked by the vision encoder (e.g., BEVFormer). The results in Appendix~\ref{lab:infer}.

\section{Conclusion}
We propose \textbf{\textit{Raw2Drive}}, the first end-to-end model-based reinforcement learning method in autonomous driving. Our approach introduces a novel dual-stream architecture and a guidance mechanism to effectively enable MBRL learning. The approach is fulfilled by the careful treatment of the two world models for the raw and privileged information, respectively, and achieve the state-of-the-art performance on lately released benchmarks. We hope this work serves as a stepping stone toward exploring reinforcement learning for end-to-end autonomous driving.

\label{sup:limitations}
\textbf{Limitations:} In our setting, the privileged input is ground truth bounding boxes and HD-Map. And in real-world autonomous driving, for industry, the ground truth bounding boxes and HD-Map can be from human annotation or advanced perception algorithms. Reinforcement learning in the real world is also a technical issue that would be solved by 3DGS~\cite{Ma2025BezierGS, huang2023iddr, chen2025focused} or a diffusion-based simulator~\cite{yang2025resim} in the future. While CARLA remains the only viable closed-loop simulator for RL research at present, our work focuses on policy learning and introduces a dual-stream world model design that is conceptually decoupled from the specific simulator.

\textbf{Social Impact:} 
Raw2Drive presents an efficient reinforcement learning framework for end-to-end autonomous driving, mitigating key issues of imitation learning such as causal confusion and distribution shift. By learning robust world models from raw sensors, it enhances the safety, reliability, and generalization of autonomous driving systems. 

\newpage



\medskip

{
\small
\bibliographystyle{unsrt}
\bibliography{neurips_2025}
}

\newpage
\appendix
\label{sec:appendix}
\section{Related Works}
\label{sup:related_work}
\subsection{End-to-End Autonomous Driving Based on Imitation Learning}
Imitation Learning (IL)~\cite{chen2024end, pan2017agile, jia2024amp, yang2025trajectory, zhu2024flatfusion, lu2024activead, yang2023llm4drive, yang2025drivemoe} is a foundational method for training agents by mimicking expert behavior, especially suitable for end-to-end autonomous driving tasks. IL aims to replicate human driving actions, thus reducing system complexity and simplifying the decision-making process from perception to control. 

Some approaches~\cite{wu2022trajectoryguided, jia2023thinktwice, jia2023driveadapter, jia2023hdgt} typically involve collecting large amounts of human driving data and using supervised learning to directly map sensor inputs (such as camera, LiDAR data, etc.) to control outputs (such as steering angle, throttle, and brake). The expert driving actions are recorded as state-action pairs, and imitation learning is used to train a model that approximates this behavior. The classic TCP model~\cite{wu2022trajectoryguided} uses a single camera input in a dual-branch design that integrates trajectory prediction and multi-step control within a single framework. It flexibly combines the outputs of both branches based on prior planning to optimize the final control signal. The latest UniAD~\cite{hu2023planning} approach goes further by using surround-view camera inputs and leveraging the query feature of transformer architectures, integrating detection, tracking, mapping, trajectory prediction, occupancy grid prediction, and planning into a single differentiable end-to-end pipeline. This design sets a new benchmark for integrating tasks in autonomous driving. To accelerate grid-based scene representation in UniAD, VAD~\cite{jiang2023vad} proposes a fully vectorized end-to-end framework that uses vectorized representations for agent motion and map elements, significantly improving task efficiency.

Although behavioral cloning methods perform well in simple driving environments, they still cannot successfully complete long-term sequence decisions in complex traffic scenarios~\cite{jia2024bench, zhai2023rethinking, you2024bench2drive}, which also called corner case. The reason is that its network uses regression loss to generate future trajectories, and scenes with large turns will be averaged by straight road scenes. For example, the recent closed-loop Bench2Drive~\cite{jia2024bench} evaluation showed that supervised learning models are sensitive to shifts in data distribution; they perform well in simple driving interactions but tend to accumulate errors in dynamic traffic settings, causing the model to deviate from the intended driving path.

\subsection{End-to-End Autonomous Driving based on Model-free Reinforcement Learning}
Model-free reinforcement learning~\cite{zhang2021end, toromanoff2020end, NEURIPS2022_8be9c134, Bai_Zhang_Tao_Wu_Wang_Xu_2023, bai2024efficient, bai2025retrieval, zhang2025amulet, li2023normalization, zhang2023gobigger, zhu2025flowrl} is another key approach for end-to-end autonomous driving, optimizing decision-making through rewards rather than direct imitation. Early RL work, such as MaRLn~\cite{toromanoff2020end}, demonstrated the potential of reinforcement learning in autonomous driving. However, it required extensive pre-training, with around 20M (23 days) of training in a single town or 50M (57 days) in multi-towns to solve only 4 standard cases in CoRL 2017~\cite{dosovitskiy2017carla}. The complexity of perception in autonomous driving makes it difficult to optimize perception and control jointly, increasing the challenge of training end-to-end systems with RL. To reduce the impact of perception on decision training, some classic RL methods, such as Roach~\cite{zhang2021end}, use 2D Bird’s Eye View (BEV) representations as input observations, simplifying the task of learning from raw sensor data and focusing on high-level environmental features. And then use RL methods to collect offline data for training the end-to-end method~\cite{wu2022trajectoryguided, shao2023safety, shao2023reasonnet}. 

\subsection{End-to-End Autonomous Driving based on Model-based Reinforcement Learning}
Model-free RL suffers challenges in complex driving scenarios or corner cases that require long-term memory. Think2Drive~\cite{li2024think} combines Roach’s~\cite{zhang2021end} 2D BEV inputs with a world model, significantly improving sampling efficiency and effectively modeling sequential interactions, leading to strong performance in complex traffic scenarios. To our best knowledge, there is no model-based RL end-to-end method in autonomous driving.

\section{Details of Training Pipeline}
\label{sec:train_pipeline}
As shown in Alogrithm~\ref{algo:overall},the training pipeline of Raw2Drive consists of two stages. In Stage I, we use privileged observations for MBRL training. The world model and behavior policy are updated alternately.

In Stage II, we train the raw sensor world with the help of the proposed guidance mechanism. Compared to Stage I, the world model is trained with an additional loss derived from the rollout guidance. For behavior policy training, we fine-tune the privileged policy by interacting with the raw sensor world model with head guidance.
\begin{algorithm}[!]
\caption{Training Pipeline of Raw2Drive}
\label{algo:overall}
\small
\NoLabelLine{Privileged Observation $o$, World Model $\textnormal{WM}$,  Policy $\pi$}
\NoLabelLine{Raw Sensor Observation $\hat{o}$, World Model $\hat{\textnormal{WM}}$,  Policy $\hat{\pi}$}
\NoLabelLine{Privileged Replay Buffer $B$,  Replay Buffer $\hat{B}$ Training iterations $N$}
\NoLabelLine{\textbf{Stage 1: Privileged World Model and Policy Training}}
\For{$i = 1$ to $N$}{
    \State Collect trajectories $(o_t, a_t, r_t, c_t, o_{t+1})$ with current $\pi$ to interact with simulator and store them in Buffer $B$\;
    
    \State Sample a trajectory $(o_{t:T}, a_{t:T}, r_{t:T}, c_{t:T}, o_{t+1:T+1})$ from Privileged Replay Buffer $B$\;
    
    \State Train Priviledged World Model $\textnormal{WM}$ with prediction loss $\mathcal{L}_{pred}$,  dynamics loss $\mathcal{L}_{dyn}$, and  representation loss $\mathcal{L}_{rep}$ with weights  $\beta_{pred}$, $\beta_{dyn}$, $\beta_{rep}$\;
    
    \State $\mathcal{L}_{\text{WM}} = \mathbb{E} \Bigg[\sum_{t=1}^{T} \big(\beta_{\text{pred}} \mathcal{L}_{\text{pred}} + \beta_{\text{dyn}}\mathcal{L}_{\text{dyn}} + \beta_{\text{rep}}\mathcal{L}_{\text{rep}} \big) \Bigg]$\;
    
    \State Sample a trajectory $o_{t:T}$ from Privileged Replay Buffer $B$\;
    
    \State Rollout in Priviledged World Model to obtain the predicted latent state $(h_{t:T}, s_{t:T})$, as well as the predicted reward and continuous flag. These predictions are then combined to construct the trajectory: $(o_{t:T}, a_{t:T}, r_{t:T}, c_{t:T}, h_{t+1:T+1}, s_{t+1:T+1})$\;
    
    \State Train privileged policy $\pi$ based on the actor-critic algorithm with the above trajectory which rollouts in Priviledged World Model\;
}
\hrulefill \\
\NoLabelLine{\textbf{Stage 2: Raw Sensor World Model and Policy Training}}
\For{$i = 1$ to $N$}{
    \State Collect trajectories $(o_t, \hat{o_t}, \hat{a}_t, \hat{r}_t, \hat{c}_t, \hat{o}_{t+1})$ by using raw sensor policy $\hat{\pi}$ to interact with simulator and store them in Raw Replay Buffer $\hat{B}$\;
    
    \State Sample a trajectory $(o_{t:t+T}, \hat{o}_{t:t+T}, \hat{a}_{t:t+T}, \hat{r}_{t:t+T}, \hat{c}_{t:t+T}, \hat{o}_{t+1:t+T+1})$ from Raw Replay Buffer $\hat{B}$\;
    
    \State Train Raw Sensor World Model $\hat{\textnormal{WM}}$ with additional loss in \textbf{\textit{Rollout Guidance}} for latent state alignment\;

    \State $\mathcal{L}_{\hat{\textnormal{WM}}} = \mathcal{L}_{\textnormal{WM}} + \mathcal{L}_{\textnormal{Rollout}}$\;
    \State Sample a trajectory $\hat{o}_{t:T}$ from Raw Replay Buffer $\hat{B}$;
    
    \State Rollout in Raw Sensor World Model to obtain the predicted latent state $(\hat{h}_{t:T}, \hat{s}_{t:T})$. Using \textbf{\textit{Head Guidance}} to obtain the predicted reward and continuous flag. These predictions are then combined to construct the trajectory: $(\hat{o}_{t:T}, \hat{a}_{t:T}, \hat{r}_{t:T}, \hat{c}_{t:T}, \hat{h}_{t+1:T+1}, \hat{s}_{t+1:T+1})$\;
    
    \State Train Raw Sensor Policy $\hat{\pi}$ based on the actor-critic algorithm with the above trajectory which rollouts in Raw Sensor World Model\;
}
\end{algorithm}

\section{Details of the Experiment}
\label{sec:exp}

\subsection{Details of the Dual Stream Input}
\label{sec:dual_stream}
For the privileged inputs, we utilize BEV semantic segmentation masks $\in \{0, 1\}^{H \times W \times C}$ as image input and ego vehicle info as vector input. Each channel in the BEV masks represents the presence of a specific type of object. It is generated from the privileged information obtained from the simulator and consists of $C$ masks of size $H \times W$. Note that the $C=43$ channels of semantic segmentation masks correspond to static objects (e.g. roads, lines, lanes, and the ego vehicle) and dynamic objects (signs, lights, pedestrians, vehicles, and obstacles). 

For the raw sensor inputs, we use BEVFormer~\cite{li2022bevformer} as the encoder surrounding RGB images to achieve grid-shape BEV features.

\subsection{Action Space in Raw2Drive}
To simplify the action space, we design a set of 39 discrete actions. The full list of actions is provided in Table~\ref{tab:actions}.

\begin{table}[H]
  \centering
  \small
  \caption{\textbf{The Discrete Actions.} The continuous action space is decomposed into 39 discrete actions, each for specific values of throttle, steer, brake and reverse. Each action is rational and legitimate.} \label{tab:actions}
  \resizebox{\linewidth}{!}{
  \begin{tabular}{cccc|cccc|cccc}
     \toprule
     Throttle & Brake & Steer & Reverse & Throttle & Brake & Steer & Reverse & Throttle & Brake & Steer & Reverse \\
    \hline
     0   & 0 &    1 & False & 0.3 & 0 & -0.2 & False & 0 & 0 & 0.1 & False \\
     0.7 & 0 & -0.5 & False & 0.3 & 0 & -0.1 & False & 0 & 0 & 0.3 & False \\
     0.7 & 0 & -0.3 & False & 0.3 & 0 & 0    & False & 0 & 0 & 0.6 & False \\
     0.7 & 0 & -0.2 & False & 0.3 & 0 & 0.1  & False & 0 & 0 & 1.0 & False \\
     0.7 & 0 & -0.1 & False & 0.3 & 0 & 0.2  & False & 0.5 & 0 & -0.5 & True \\
     0.7 & 0 & 0    & False & 0.3 & 0 & 0.3  & False & 0.5 & 0 & -0.3 & True \\
     0.7 & 0 & 0.1  & False & 0.3 & 0 & 0.5  & False & 0.5 & 0 & -0.2 & True \\
     0.7 & 0 & 0.2  & False & 0.3 & 0 & 0.7  & False & 0.5 & 0 & -0.1 & True \\
     0.7 & 0 & 0.3  & False & 0   & 0 & -1   & False & 0.5 & 0 & 0 & True \\
     0.7 & 0 & 0.5  & False & 0   & 0 & -0.6 & False & 0.5 & 0 & 0.1 & True \\
     0.3 & 0 & -0.7 & False & 0   & 0 & -0.3 & False & 0.5 & 0 & 0.2 & True \\
     0.3 & 0 & -0.5 & False & 0   & 0 & -0.1 & False & 0.5 & 0 & 0.3 & True \\
     0.3 & 0 & -0.3 & False & 1   & 0 & 0    & False & 0.5 & 0 & 0.5 & True \\
    \bottomrule
  \end{tabular}
  }
\end{table}

\subsection{Model Configuration}
We implement the model using PyTorch. Both world models are trained with a learning rate of 1e-5, weight decay of 0.00, and the AdamW optimizer. The behavior policy is trained with a learning rate of 3e-5, weight decay of 0.00, also using AdamW. The weights $\beta_e$, $\beta_h$, and $\beta_s$ are set to 10, 5, and 10, respectively.

\subsection{Reward Design in Raw2Drive}
We adopt the reward design and reward shaping approach from Think2Drive~\cite{li2024think}.

\section{Details of the losses in World Model}
\label{sec:loss_wm}
Train the World Model $W_p$ using prediction loss $\mathcal{L}_{pred}$, the dynamics loss $\mathcal{L}_{dyn}$, and the representation loss $\mathcal{L}_{rep}$, the loss weights are respectively $\beta_{pred}$, $\beta_{dyn}$, $\beta_{rep}$: \\
\begin{equation}
\begin{aligned}
\mathcal{L}_{\text{pred}}(\theta) \doteq& - \ln p_{\theta} (e_t \mid s_t, h_t) - \ln p_{\theta} (r_t \mid s_t, h_t) \\ 
& - \ln p_{\theta} (c_t \mid s_t, h_t) \\ 
\mathcal{L}_{\text{dyn}}(\theta) \doteq& \max \left( 1, \text{KL} \left[ \text{sg} \left( q_{\theta} (s_t \mid h_t, e_t) \right) \parallel p_{\theta} (s_t \mid h_t) \right] \right) \\
\mathcal{L}_{\text{rep}}(\theta) \doteq& \max \left( 1, \text{KL} \left[ q_{\theta} (s_t \mid h_t, e_t) \parallel \text{sg} \left( p_{\theta} (s_t \mid h_t) \right) \right] \right) \\
\end{aligned}
\end{equation}

\section{Details of the losses in Behavior Policy}
\label{sec:loss_bp}
The critic network $p_\psi$ estimates the distribution of future returns $R^\lambda_t$, while the value function $v_t$ represents the expected value of the return at state $s_t$, train Critic network in the Privileged Behavior Policy $\pi_p$ by maximum likelihood loss:
\begin{equation}
\begin{aligned}
        \mathcal{L}(\psi) \, \dot{=}& - \sum_{t=1}^{T} \ln p_{\psi}(R_t^\lambda | s_t) \\
        R_t^\lambda \, \dot{=}& r_t + \gamma c_t \left( (1 - \lambda)v_t + \lambda R_{t+1}^{\lambda} \right) \\
        R_T^\lambda \, \dot{=}& v_T
\end{aligned}
\end{equation}
Train the actor network through policy optimization, using entropy regularization $\mathcal{H}$, exponential moving average (EMA) for smoothing, entropy regularization coefficient $\eta$ and the \text{sg} operation for gradient stability.
\begin{equation}
\begin{aligned}
    \mathcal{L}(\theta) \, \dot{=}& - \sum_{t=1}^{T} \, \text{sg} \left( \frac{(R_t^\lambda - v_{\psi}(s_t))}{\max(1, S)} \right) \log \pi_{\theta}(a_t | s_t) \notag \\
    &+ \eta \mathcal{H} \left[ \pi_{\theta}(a_t | s_t) \right] \\
    S \, \dot{=}& \text{EMA} \left( \text{Per}(R_t^\lambda, 95) - \text{Per}(R_t^\lambda, 5), 0.99 \right)
\end{aligned}
\end{equation}

\section{Rollout Guidance}
\label{sup:rollout_guidance}
\begin{figure}[!tb]
    \centering
    \includegraphics[width=0.75\textwidth]{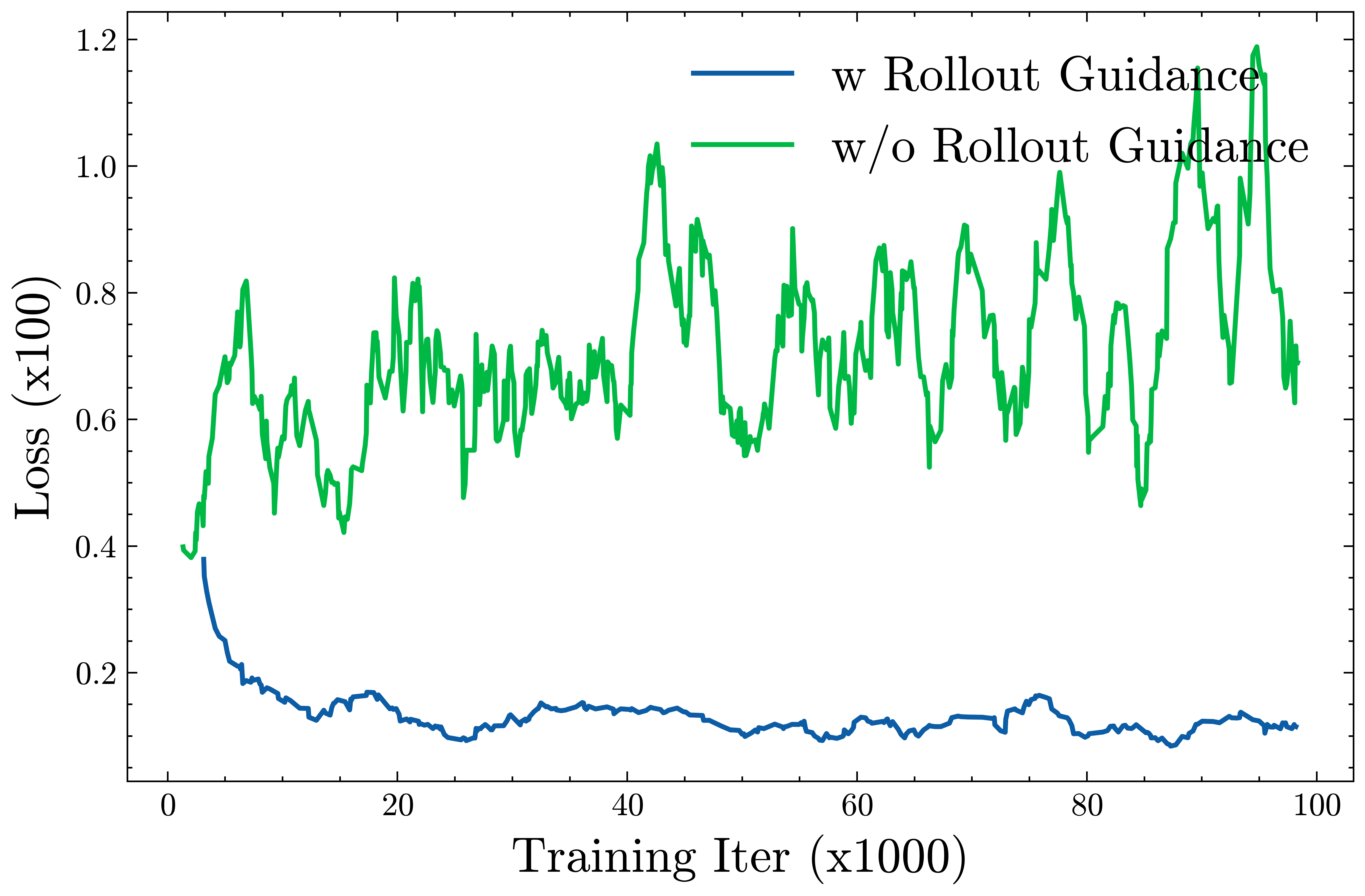}
    \caption{\textbf{Comparison of the Raw Sensor World Model Loss w/ and w/o Guidance}.}
    \label{fig:loss_guidance}
\end{figure}

As shown in Figure~\ref{fig:loss_guidance}, the rollout guidance ensures consistency between the dual-stream world models during rollouts, playing a critical role in training the raw sensor world model; without it, the network struggles to converge.

\section{Inference Latency}
\label{lab:infer}
\begin{table}[h]
\centering
\caption{\textbf{Latency Comparison between Privileged and Raw Sensor Streams.}}
\resizebox{\linewidth}{!}{
\begin{tabular}{l l c c c}
\toprule
\textbf{Method} & \textbf{Modality} & \textbf{Encoder Latency (ms)} & \textbf{World Model Latency (ms)} & \textbf{Policy Latency (ms)} \\
\midrule
Privileged Stream & BEV State & 2 (5$\times$Conv) & 2 & 2 \\
Raw Sensor Stream & Multi-view images & 600 (BEVFormer) & 2 & 2 \\
\bottomrule
\end{tabular}}
\end{table}

\section{Visualization}
\subsection{Decoder Output by the World Model}
\begin{figure}[!tb]
    \centering
    \includegraphics[width=1\linewidth]{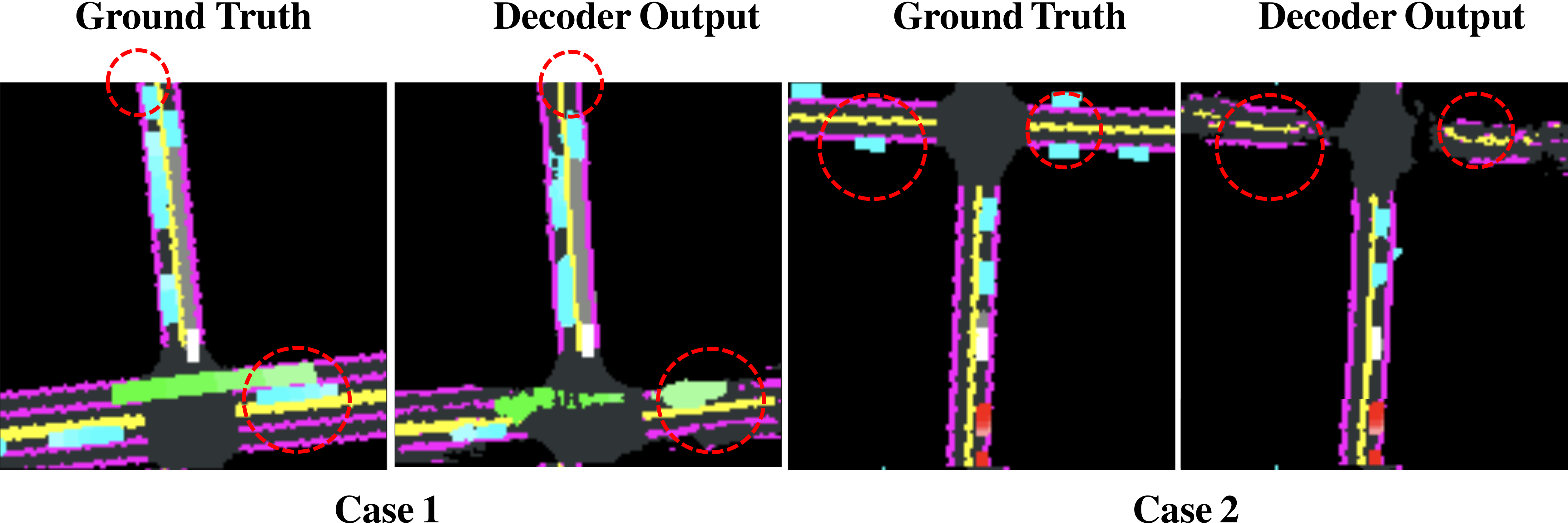}
    \caption{\textbf{Ground Truth VS Raw Sensor Decoder Output by the World Model} (The red circle is the blind spot of the camera).}
    \label{fig:gaussion_vis}
\end{figure}

We visualize the output of the raw sensor decoder to evaluate its reconstruction quality. In our dual-stream architecture, the raw sensor world model replaces expensive video reconstruction with a more efficient BEV reconstruction. As shown in the Figure~\ref{fig:gaussion_vis}, under the guidance mechanism, the raw sensor world model successfully learns to reconstruct the BEV representation with reasonable accuracy. However, due to the inherent limitations of the camera sensor, the model struggles to reconstruct occluded regions or areas beyond its direct line of sight.

\subsection{Perception Confusion}
\label{sup:perception_confuse}
\begin{figure}[!]
    \centering
    \includegraphics[width=1\linewidth]{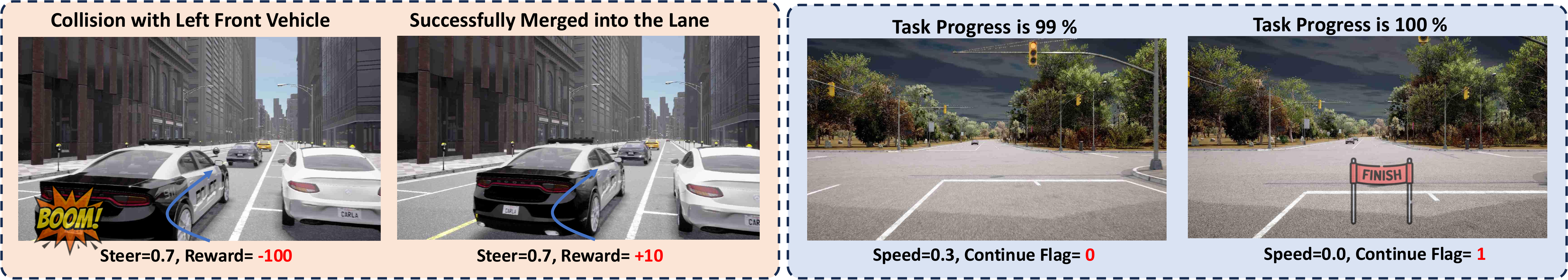}
    \caption{\textbf{Perception Confusion in Reward and Continuous Head}.
    }
    \label{fig:reward_cont}
\end{figure}

As shown in Figure~\ref{fig:reward_cont}, adjacent frames exhibit a high degree of similarity. In the upper section, the ego vehicle is in a corner case of \textbf{\textit{ParkingExit}}. Despite minimal visual differences between adjacent frames, identical actions can yield opposite rewards, causing model confusion. In the lower section, as the ego vehicle nears task completion, the visual change remains subtle. However, the binary nature of the continuation flag (0 or 1) provides limited information, further hindering convergence.

\end{document}